\documentclass[runningheads]{llncs}

\usepackage[T1]{fontenc}
\usepackage[utf8]{inputenc}
\usepackage{hyperref}
\usepackage{graphicx} 
\usepackage{float}
\usepackage{tabularx}
\usepackage{booktabs}
\usepackage{multirow}
\usepackage{amsmath}
\usepackage{cite}
\usepackage{pgfplots}

\usepackage{array} 

\definecolor{seaborn_blue}{RGB}{76, 114, 176}
\definecolor{seaborn_orange}{RGB}{221, 132, 82}

\title{Autonomous End-to-End SOH Prediction Services for Battery Systems via Temporal-Contrastive Representation Learning}
\titlerunning{TC-SOH: Temporal-Contrastive SOH Prediction}

\author{Junting Wen\inst{1} \and
        Dan Li\inst{1} \thanks{Corresponding author.} \and
        Qihao Quan\inst{1} \and
        Xiwen Wang\inst{1} \and
        Hang Yang\inst{1} \and
        Zhaohong Meng\inst{1} \and
        Zigui Jiang\inst{1} \and
        Changlin Yang\inst{2} \and
        Tianle Liu\inst{3} \and
        Diego Muñoz-Carpintero\inst{5} \and
        Jian Lou\inst{1}}

\authorrunning{J. Wen et al.}
\institute{
    School of Software Engineering, Sun Yat-sen University, Zhuhai 519082, China\\
    \email{\{lidan263, louj5, jiangzg3\}@mail.sysu.edu.cn} \and
    Tianneng Battery Group Co., Ltd, Zhejiang, China\\
    \email{yangcl02@tn-ess.com} \and
    School of Communication Engineering, Hangzhou Dianzi University, China\\
    \email{tianle@hdu.edu.cn}\and
    Institute of Engineering Science, Universidad de O'Higgins, Rancagua, Chile\\
    \email{diego.munoz@uoh.cl}
}

\begin{document}

\flushbottom
\maketitle
\begin{abstract}
Accurate state of health (SOH) estimation is a critical diagnostic service for lithium-ion battery management. However, reliance on labor-intensive manual feature engineering and opaque black-box models hinders scalable industrial deployment. To address this, we introduce TC-SOH: a modular, plug-and-play service architecture for autonomous, end-to-end SOH prediction. TC-SOH employs a temporal-contrastive mechanism and a cross-window prediction pretext task to extract degradation-relevant representations directly from raw operational data. To improve transparency, we connect model efficacy with representation diagnostics: visualization, sensitivity analysis, redundancy analysis, bidirectional probing, future-SOH probing, and temporal shuffling show that learned features overlap with selected expert descriptors while retaining additional SOH-relevant variation, and that ordered temporal context improves subsequent-SOH prediction. Across four public datasets, TC-SOH outperforms the considered physics-informed and data-driven baselines, reducing MAPE by 1.91 times and RMSE by 2.13 times.

\end{abstract}

\section{Introduction}

In modern energy services, lithium-ion batteries (LIBs) function as the foundational physical assets for electric mobility, portable electronics, and grid-scale energy storage. Their ubiquitous deployment is driven by their capacity to deliver high-quality energy services, characterized by superior energy density, power density, and extended lifecycle durability \cite{ZAGHIB20113949}.

Driven by the demand for uninterrupted energy services, the lithium-ion battery market is projected to reach USD 221.16 billion by 2030 (12.19\% CAGR)\cite{Markets2024}, necessitating robust, lifecycle-spanning asset management. However, maintaining consistent service delivery is challenged by progressive asset degradation. As complex aging mechanisms compromise service reliability \cite{LI2025116078} and operational safety \cite{adebanjo2025comprehensive}, developing highly reliable battery health diagnostic services is imperative, which empowers modern Battery Management Systems (BMS) to proactively mitigate risks and maximize the lifecycle value of these energy assets.

Within this service-oriented framework, SOH serves as the core quantitative metric for diagnosis, defined by the ratio of current to initial capacity. Typically, an 80\% SOH marks the asset's retirement from its primary role \cite{XIA2023107161}. Although full charge-discharge cycles offer accurate capacity measurement via ampere-hour integration \cite{CAO2025125086}, such tests cause unacceptable service disruptions, induce further asset degradation \cite{BARRE2013680}, and are impractical in dynamic environments. Therefore, to meet the industrial demand for continuous, automated SOH diagnostics, modern management protocols must extract actionable insights from partial routine data, entirely avoiding disruptive complete charge-discharge profiles.

While data-driven methodologies serve as powerful computational engines for modern SOH diagnostic services \cite{REN2025124385,NI2025125539}, their scalable industrial deployment is fundamentally bottlenecked by two critical barriers. First, the efficacy of these service architectures relies heavily on labor-intensive, bespoke feature engineering \cite{XU2025136155}. Because indirect health indicators—such as state of charge (SOC), thermal variations, and ohmic resistance \cite{S2025100237,ma18010145,WU2018370,KHODADADISADABADI2021228861}—are highly sensitive to fluctuating operational conditions, conventional extraction methods inevitably discard subtle, non-linear degradation dynamics \cite{ZENG2025236608,MU2025134578}, while arbitrarily expanding feature sets merely inflates computational overhead \cite{XIONG2023460}. Second, the opaque, "black-box" nature of deep learning models creates a profound trust deficit between the diagnostic service and industrial end-users. When anomalies occur, the inability to trace data-driven predictions back to physical electrochemical root causes compromises service transparency and reliability. Ultimately, this lack of interpretability severely limits the practical adoption of these AI-driven diagnostic services in rigorous, safety-critical asset management environments.

Representation Learning \cite{bengio2014representationlearningreviewnew} enhances service scalability by autonomously distilling compact embeddings from raw operational data, decoupling SOH diagnostics from rigid, manual priors \cite{XIANG2024101763}. Crucially, deep architectures capture the subtle, non-linear degradation dynamics that conventional extraction methods frequently discard as transient noise. This automated approach isolates invariant health features that remain robust across fluctuating environmental conditions, while simultaneously mitigating the computational overhead that typically plagues arbitrarily expanded, hand-crafted feature sets. However, despite achieving high predictive accuracy, deep architectures like autoencoders \cite{JIANG2023232466} produce opaque latent spaces. This lack of geometric interpretation regarding degradation trajectories destroys transparency, severely limiting deployment in safety-critical asset management \cite{blackbox, YE2024130828}.

To rebuild end-user trust in safety-critical environments, an SOH service should not only predict accurately but also explain what its representation contributes. Unlike standard physics-informed models\cite{krishnapriyan2021characterizingpossiblefailuremodes}, our analysis explicitly diagnoses the learned representation: whether it tracks smooth degradation trajectories, remains sensitive in early SOH regimes, overlaps with physical descriptors without being exhausted by them, and whether its temporal context predicts subsequent SOH only when cycle order is preserved. This diagnostic view aligns model capability with the reliability requirements of industrial asset management.

Addressing the urgent need for scalable asset management, we introduce TC-SOH: a cross-disciplinary plug-and-play diagnostic service architecture. Operating as an end-to-end, modular service plug-in, TC-SOH integrates into existing BMS frameworks while reducing labor-intensive manual feature engineering. By combining temporal-contrastive learning with cross-window prediction proxy tasks, the module extracts non-linear aging evolution from raw operational data and summarizes ordered cycle history into temporal context. We further evaluate this context through a diagnostic chain of visualization, sensitivity, redundancy, probing, future-SOH probing, and temporal shuffling, thereby linking predictive efficacy to interpretable representation behavior. The main contributions are summarized as follows:

\begin{itemize} 
\item \textbf{Plug-and-Play Diagnostic Framework:} We propose an end-to-end TC-SOH service framework that uses temporal-contrastive learning and cross-window prediction to extract non-linear aging dynamics from routine CC-CV data, reducing reliance on manual feature engineering.  
\item \textbf{Service Trust and Transparency:} We establish a diagnostic interpretation chain combining trajectory visualization, early-stage sensitivity checks, redundancy analysis, bidirectional probing, future-SOH probing, and temporal shuffling to show how learned features relate to physical descriptors and how TC context contributes ordered-history information.
\item \textbf{Service Efficacy and Robustness:} Validation on four large-scale datasets confirms TC-SOH's operational efficacy and transferability. It outperforms the considered data-driven and physics-hybrid baselines under the repeated battery-level protocol, reducing MAPE by 1.91x and RMSE by 2.13x. \end{itemize}

\section{Data description }
We evaluate TC-SOH on four widely used public benchmarks (XJTU \cite{wang_physics-informed_2024}, TJU \cite{Zhu2022}, HUST \cite{Ma2022}, and MIT \cite{Severson2019}), covering 386 batteries with diverse chemistries, operating temperatures, and charging/discharging protocols. Table~\ref{tab:battery_information} summarizes the main dataset attributes, while the text below highlights the complementary role of each benchmark.




\begin{table}[h]
\vspace{-3mm}
\small
\centering
\renewcommand{\arraystretch}{1.4}
\resizebox{\textwidth}{!}{%
\begin{tabular}{l | c c c c c c}
    \toprule\midrule
    \textbf{Dataset} & \textbf{HUST} & \textbf{MIT} & \textbf{XJTU} & \textbf{TJU (Batch 1)} & \textbf{TJU (Batch 2)} & \textbf{TJU (Batch 3)} \\
    \midrule
    Subset & 1-10 & 1-3 & 1--6 & 1 & 2 & 3 \\
    Nominal capacity (mAh) & 1100 & 1100 & 2000 & 3500 & 3500 & 2500 \\
    Cut-off voltage (V) & 2.0--3.6 & 2.0--3.6 & 2.5--4.2 & 2.65--4.2 & 2.5--4.2 & 2.5--4.2 \\
    Experiment temperature (°C) & 30 & 30 & Room temperature & 25,35,45 & 25,35,45 & 25 \\
    \midrule
    \bottomrule
\end{tabular}%
}
\caption{Summary of Battery Information}
\label{tab:battery_information}
\vspace{-9mm}
\end{table}

\noindent\textbf{MIT and HUST.}
MIT and HUST form a controlled pair of LFP benchmarks because both use the same A123 cell model with the same nominal capacity. MIT emphasizes charging diversity through one-step and two-step fast-charging policies, whereas HUST keeps the environmental setting fixed at 30$^\circ$C and varies the discharge conditions. This pairing helps separate charging-driven variation from discharge-driven variation.

\smallskip
\noindent\textbf{XJTU.}
XJTU contains LISHEN NCM523 cells cycled at room temperature under six operating subsets. The charging protocol is fixed for the first five subsets, while the discharge side covers fixed-current, random-current, and satellite GEO walking profiles. This design makes XJTU a protocol-rich benchmark for testing whether the learned representation remains stable under changing usage patterns.

\smallskip
\noindent\textbf{TJU.}
TJU is the most heterogeneous dataset in our study, spanning NCA, NCM, and NCM+NCA cells, temperatures from 25 to 45$^\circ$C, and C-rates from 0.25 to 4.0 C. Its broad chemistry and operating range provides a strong stress test for robustness across cross-domain conditions.
\vspace{-2mm}
\section{Methods}
\vspace{-2mm}
This section describes the data processing workflow, the battery degradation process, and the TC-SOH framework, detailing the design of each module and the data augmentation techniques.
\vspace{-2mm}
\subsection{Selection of CC-CV phase} Battery aging is driven by irreversible side reactions, including electrode material loss and cathode structural degradation \cite{TANG2025126870}. Although a full cycle comprises CC charging, CV charging, relaxation, and discharge, we specifically utilize raw data from the CC-CV stage for feature extraction. This selection is motivated by three factors: First, the CC-CV stage provides the highest contribution to SOH prediction accuracy in the experiment of \cite{LYU2023129067}. Second, this stage are readily accessible and widely recorded in real-world applications, ensuring the framework's practical utility. Finally, selecting data prior to full charge minimizes the influence of variable cycling conditions and external factors \cite{wang_physics-informed_2024}, which is critical for ensuring that learned features represent intrinsic battery characteristics rather than specific experimental protocols \cite{GESLIN20231956}. This selection enhances the model's robustness and generalizability across diverse battery chemistries and charging/discharging profiles.

\subsection{Framework}
\begin{figure*}[htbp]
    \vspace{-6mm}
    \centering
    
    \includegraphics[width=\linewidth, trim=120 170 120 130, clip]{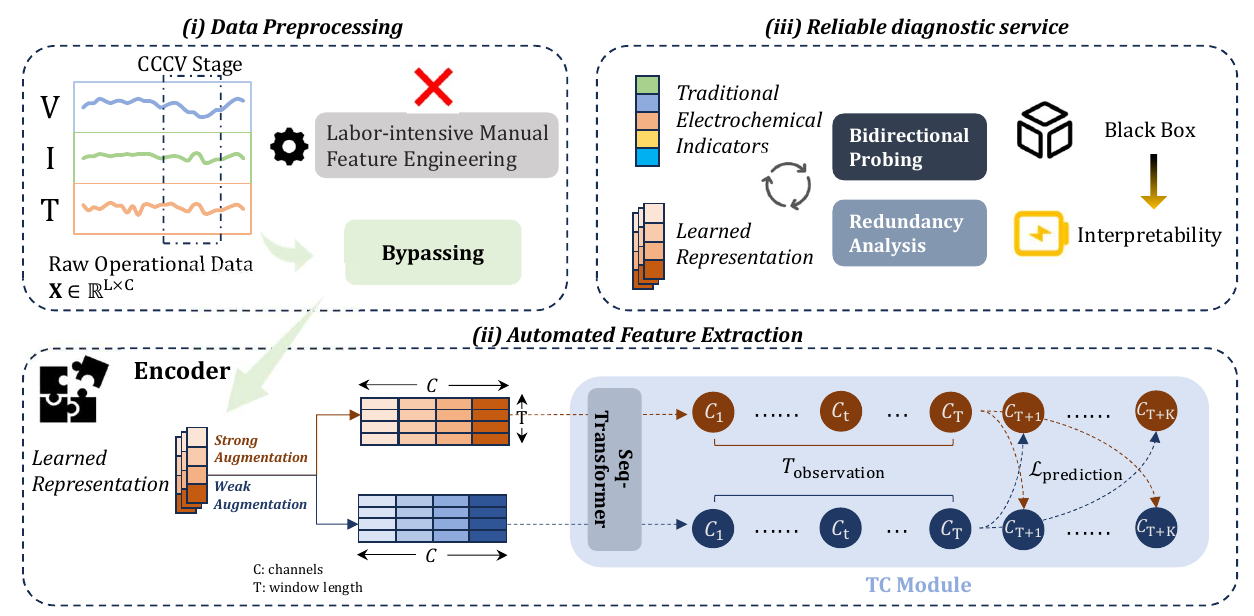}
    \vspace{-4mm}
    \caption{Overview of TC-SOH framework}
    \label{fig:framework}
    \vspace{-1cm}
\end{figure*}

The battery aging process is a multifaceted phenomenon involving complex electrochemical reactions influenced by battery chemistry, operational conditions, and degradation stages. While traditional empirical models approximate capacity fade as a univariate function of time or cycle count, they often lack generalizability across diverse protocols \cite{6228850}. Formally, we define the degradation trajectory as an estimation function $SOH = f(\mathbf{x})$, where $\mathbf{x}$ denotes an information-rich feature vector derived from raw battery data. To optimize both feature extraction and functional approximation, we propose the TC-SOH framework. As illustrated in Figure \ref{fig:framework}, the architecture comprises three integrated components:\begin{enumerate}\item \textbf{CNN Module $f(\cdot)$:} Leverages the robust data fusion capabilities of 1D-CNNs to transform raw electrical signals from individual cycles into high-dimensional latent representations.\item \textbf{TC Module $g(\cdot)$:} Enhances the discriminative power of these features by performing a cross-prediction task on augmented temporal sequences, thereby enforcing the learning of sequential degradation dynamics.\item \textbf{Output Module $o(\cdot)$:} A linear layer that maps the refined temporal representations to the final SOH value.\end{enumerate}By decoupling complex feature learning from the regression task, TC-SOH achieves superior reliability, accuracy, and generalizability in SOH estimation across heterogeneous datasets.

\subsection{Design Principles}The feature extractor $f(\cdot)$ employs a 1D-CNN architecture \cite{KIRANYAZ2021107398} to map raw electrical signals $(x_{current},x_{voltage},x_{soc})$ from a single CC-CV cycle into a high-dimensional representation $z_{c}$. The temporal window size of the convolution serves as a key tunable hyperparameter. Inspired by \cite{Eldele_2023}, the Temporal Contrastive (TC) module $g(\cdot)$ executes a cross-prediction task via a sequence transformer \cite{vaswani2023attentionneed}. It summarizes a sequence of consecutive cycle representations $z_{\leq t}$ into a context vector $c_t$ to predict subsequent representations $z_{t+1},\dots, z_{t+K}$. We model the mutual information between context $c_t$ and target $z_{t+K}$ using a log-bilinear model:$$f_{k}(z_{t+K},c_{t})=\exp{(W_{k}(c_{t})^{T} z_{t+K})},$$where $W_k$ is a linear mapping of $c_{t}$ to the dimension of $z_{t+K}$. The contrastive loss is formulated as:\begin{equation}\mathcal{L}_{TC}^s=-\frac{1}{K} \sum{k=1}^{K} \log \frac{\exp ((W_k (c_t^s))^T z_{t+k}^{w})}{\sum_{n \in N_{t, k}} \exp ((W_k (c_t^s))^T z_{n}^w)}\end{equation}\begin{equation}\mathcal{L}_{TC}^w=-\frac{1}{K} \sum{k=1}^{K} \log \frac{\exp ((W_k (c_t^w))^T z_{t+k}^{s})}{\sum_{n \in N_{t, k}} \exp ((W_k (c_t^w))^T z_{n}^s)}\end{equation}where $N_{t, k}$ denotes other samples in the mini-batch. This formulation maximizes the similarity between the predicted and true representations while minimizing similarity with negative samples, thereby enhancing the encoder's discriminative capability. The window length $\omega$ determines the span of consecutive cycles in each training sequence, directly influencing context summarization and contrastive learning.The overall training objective combines prediction and contrastive losses into a joint optimization problem:\begin{equation}\mathcal{L} = \mathcal{L}{\text{pred}} + \lambda \mathcal{L}{\text{con}},\end{equation}where $\lambda$ is empirically set to 0.6 to balance the contrastive term. This joint formulation enforces temporal consistency across cycles, effectively functioning as an auto-regressive prediction task.

\subsection{Augmentation}
We apply augmentation to the learned representations rather than raw signals to prevent encoding bias. Given the sensitivity of contrastive learning to augmentation selection \cite{chen2020simpleframeworkcontrastivelearning}, particularly for time-series \cite{Wen_2021}, we specifically adopt jitter and scaling while excluding permutation. The augmentation intensity is regulated via hyperparameters to optimize the trade-off between accuracy and generalization.

%
%
%
%
%
%
%

\section{Results and Analysis}

\subsection{Experiment Setups}
\label{subsec:setup}
To rigorously evaluate performance, we benchmark TC-SOH against the considered Physics-Informed Neural Network (PINN)\cite{wang_physics-informed_2024} and a baseline Manual-MLP trained on 16 expert-defined features from the PINN4SOH processed datasets, thereby isolating the efficacy of our automatic feature extraction against physics-guided and manual engineering approaches. PINN and Manual-MLP are evaluated under the same strict paired protocol: held-out test batteries are fixed before any window extraction, five-fold cross-validation is performed only on the remaining training/validation batteries, and the protocol is repeated over five paired seeds. The exact held-out test battery IDs for all subsets are provided in the supplementary material. Internally, we conducted ablation studies on the encoder architecture $f(\cdot)$--examining MLP, LSTM, and CNN variants--to optimize the balance between temporal dependency modeling and local feature fusion.

\begin{figure}[t] 
    \vspace{-4mm}

    \centering
    \begin{minipage}[b]{0.24\textwidth}
        \centering
        \includegraphics[width=\linewidth, trim=0 10 80 5, clip]{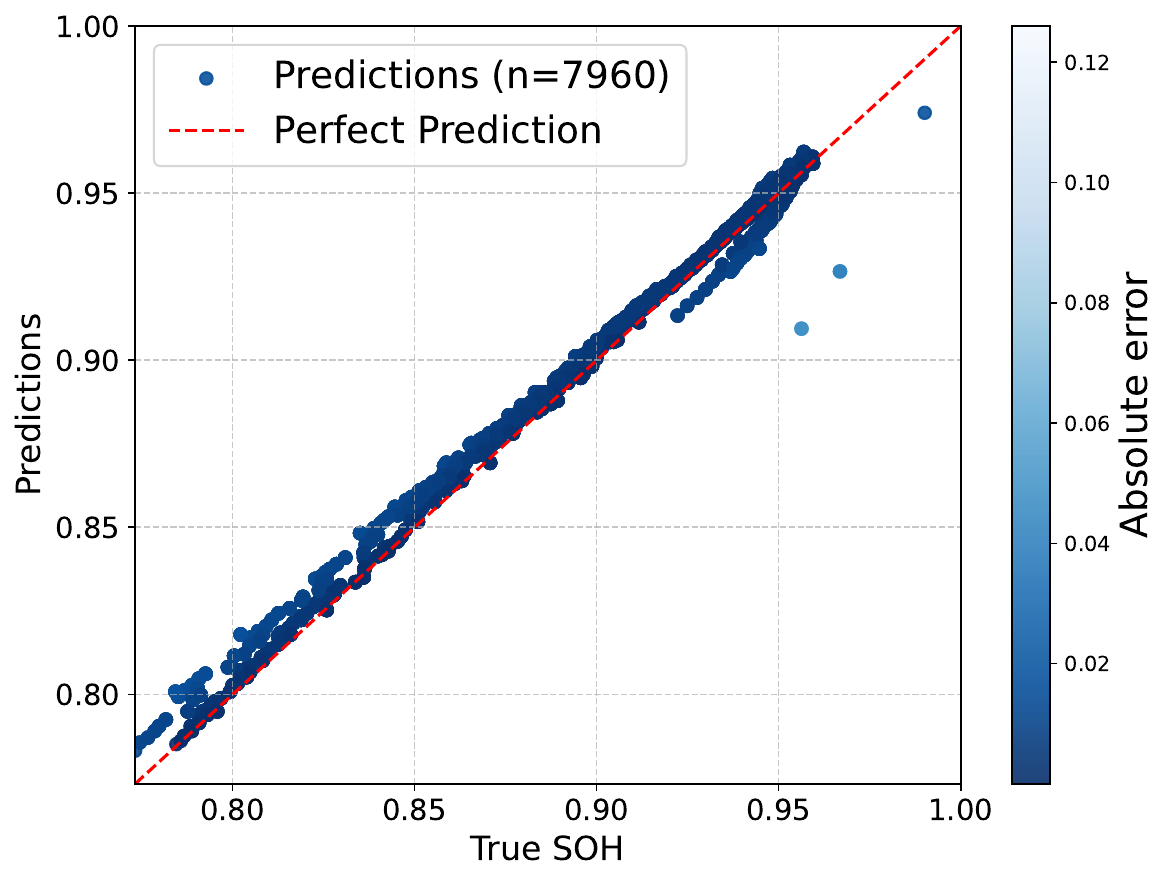}
        \vspace{-2mm}
        \centerline{\footnotesize (a) XJTU}
    \end{minipage}%
    \hfill 
    \begin{minipage}[b]{0.24\textwidth}
        \centering
        \includegraphics[width=\linewidth, trim=0 10 80 5, clip]{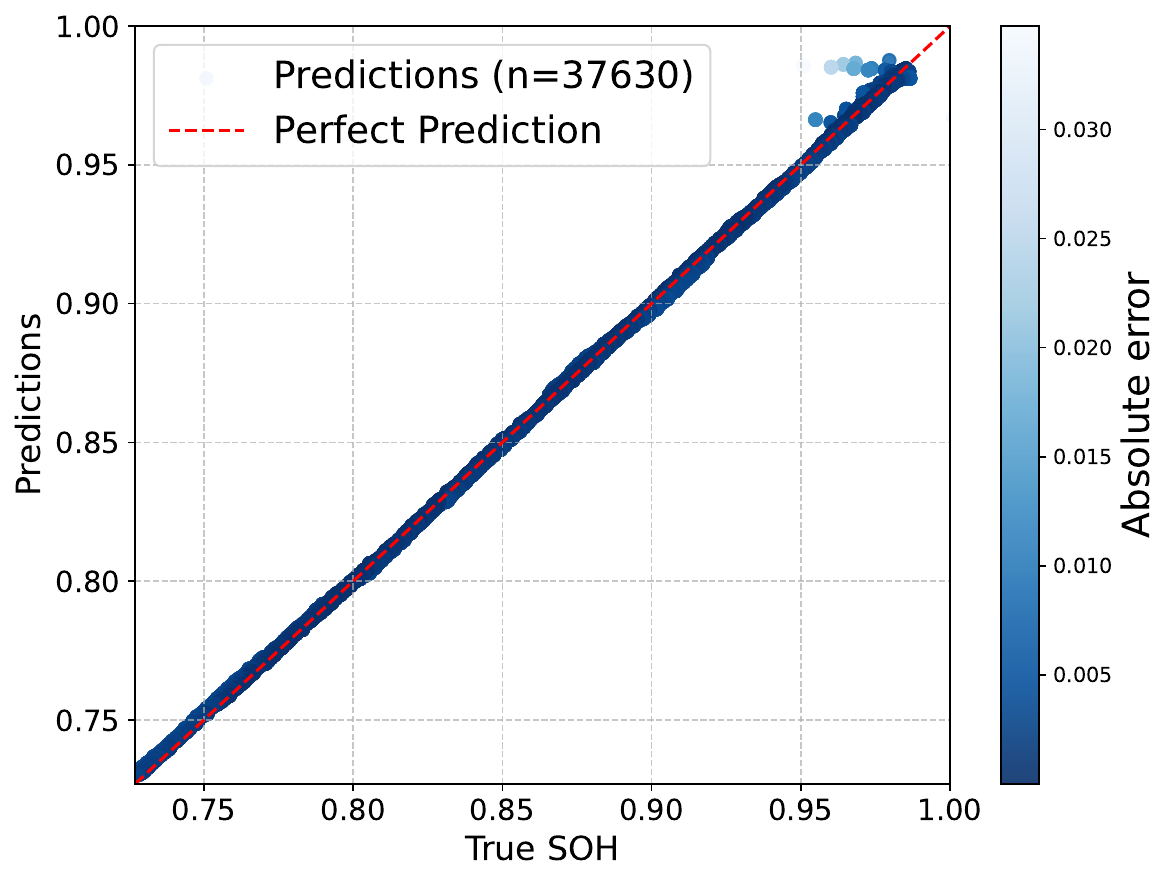}
        \vspace{-2mm}
        \centerline{\footnotesize (b) HUST}
    \end{minipage}%
    \hfill
    \begin{minipage}[b]{0.24\textwidth}
        \centering
        \includegraphics[width=\linewidth, trim=0 10 80 5, clip]{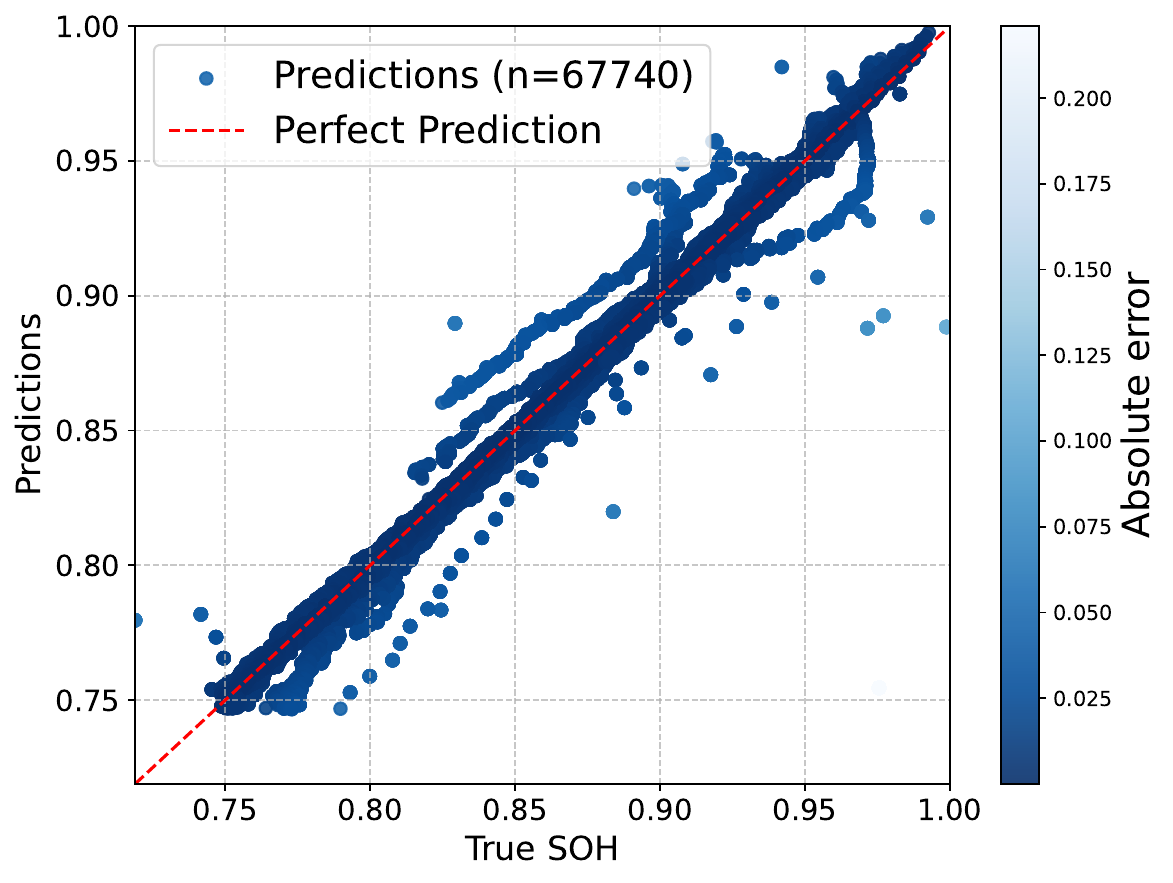}
        \vspace{-2mm}
        \centerline{\footnotesize (c) TJU}
    \end{minipage}%
    \hfill
    \begin{minipage}[b]{0.24\textwidth}
        \centering
        \includegraphics[width=\linewidth, trim=0 10 80 5, clip]{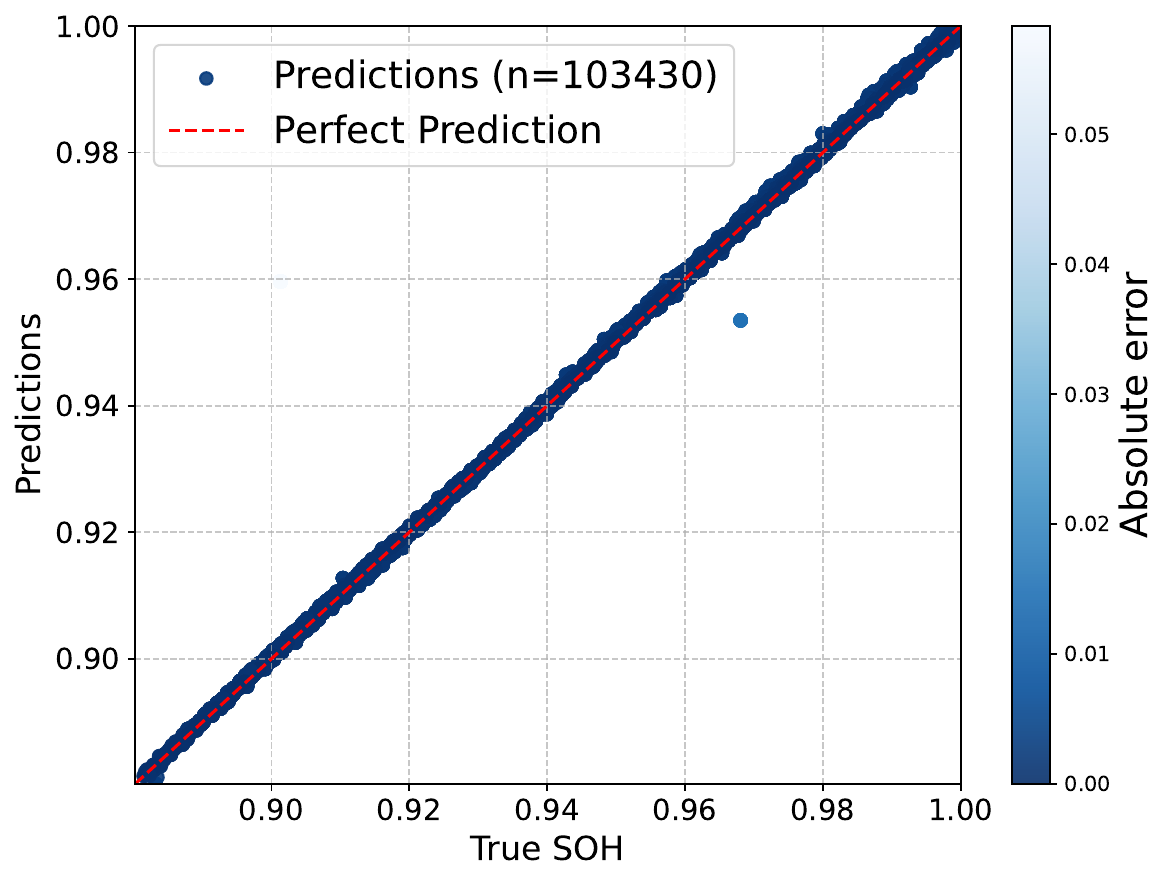}
        \vspace{-2mm}
        \centerline{\footnotesize (d) MIT}
    \end{minipage}

    \caption{
    \textbf{SOH estimation visualization.} Points near the diagonal indicate better performance.}
    \label{fig:prediction-label}
    \vspace{-2mm}
\end{figure}

\subsection{SOH Estimation}
\vspace{-2mm}
Table \ref{tab:main_results} compares the SOH estimation results of our model against the considered manual-feature and automated baselines (CNN, MLP). To avoid leakage from overlapping cycle windows, all splits are constructed at the battery level before window generation; repeated folds and random seeds are treated as matched evaluation units rather than independent window-level samples. Evaluated via RMSE and MAPE, the proposed temporal-contrastive predictors achieve the lowest mean errors on most dataset--metric pairs, including HUST with Auto-LSTM at 0.102{\scriptsize$\pm$0.020}\% MAPE and 0.0011{\scriptsize$\pm$0.0002} RMSE. The visualization of these results is shown in Fig. \ref{fig:prediction-label}. Furthermore, ablation studies confirm that the TC module significantly enhances performance by capturing intra- and inter-cycle dynamics.
\begin{table*}[t]
\centering
\caption{\textbf{Main results under repeated battery-level splits.} Values are reported as mean{\scriptsize$\pm$std} across repeated seeds after averaging the five folds and all subsets within each dataset. MAPE is given in percent; RMSE is reported in the original SOH scale. The lowest mean error in each dataset is highlighted in bold.}
\label{tab:main_results}
\footnotesize
\setlength{\tabcolsep}{4.2pt}
\renewcommand{\arraystretch}{1.08}
\begin{tabular}{lcccc}
\toprule
\textbf{Method} & \textbf{MIT} & \textbf{XJTU} & \textbf{TJU} & \textbf{HUST} \\
\midrule
\multicolumn{5}{c}{\textit{MAPE (\%)}} \\
\midrule
Auto-CNN &
0.150{\scriptsize$\pm$0.025} &
0.609{\scriptsize$\pm$0.251} &
\textbf{0.815{\scriptsize$\pm$0.016}} &
0.333{\scriptsize$\pm$0.090} \\
Auto-MLP &
0.213{\scriptsize$\pm$0.030} &
0.907{\scriptsize$\pm$0.288} &
1.376{\scriptsize$\pm$0.016} &
1.709{\scriptsize$\pm$1.688} \\
Auto-LSTM &
\textbf{0.071{\scriptsize$\pm$0.013}} &
\textbf{0.334{\scriptsize$\pm$0.002}} &
1.051{\scriptsize$\pm$0.031} &
\textbf{0.102{\scriptsize$\pm$0.020}} \\
PINN &
0.788{\scriptsize$\pm$0.029} &
1.016{\scriptsize$\pm$0.040} &
1.365{\scriptsize$\pm$0.045} &
0.868{\scriptsize$\pm$0.021} \\
Manual-MLP &
0.795{\scriptsize$\pm$0.009} &
1.266{\scriptsize$\pm$0.024} &
1.443{\scriptsize$\pm$0.065} &
1.729{\scriptsize$\pm$0.036} \\
\midrule
\multicolumn{5}{c}{\textit{RMSE}} \\
\midrule
Auto-CNN &
0.0021{\scriptsize$\pm$0.0003} &
0.0100{\scriptsize$\pm$0.0066} &
\textbf{0.0144{\scriptsize$\pm$0.0002}} &
0.0031{\scriptsize$\pm$0.0008} \\
Auto-MLP &
0.0028{\scriptsize$\pm$0.0004} &
0.0141{\scriptsize$\pm$0.0081} &
0.0188{\scriptsize$\pm$0.0001} &
0.0154{\scriptsize$\pm$0.0154} \\
Auto-LSTM &
\textbf{0.0012{\scriptsize$\pm$0.0001}} &
\textbf{0.0051{\scriptsize$\pm$0.0000}} &
0.0158{\scriptsize$\pm$0.0003} &
\textbf{0.0011{\scriptsize$\pm$0.0002}} \\
PINN &
0.0102{\scriptsize$\pm$0.0004} &
0.0124{\scriptsize$\pm$0.0005} &
0.0144{\scriptsize$\pm$0.0004} &
0.0100{\scriptsize$\pm$0.0003} \\
Manual-MLP &
0.0103{\scriptsize$\pm$0.0002} &
0.0153{\scriptsize$\pm$0.0002} &
0.0155{\scriptsize$\pm$0.0006} &
0.0226{\scriptsize$\pm$0.0005} \\
\bottomrule
\end{tabular}
\end{table*}

\subsection{Data Efficiency and Cross-Domain Adaptation}

\subsubsection{Small-Sample Efficiency}
\begin{figure}[hbtp]
    \vspace{-2mm}
    \centering
    \includegraphics[width=0.8\linewidth,trim=0 10 20 5, clip]{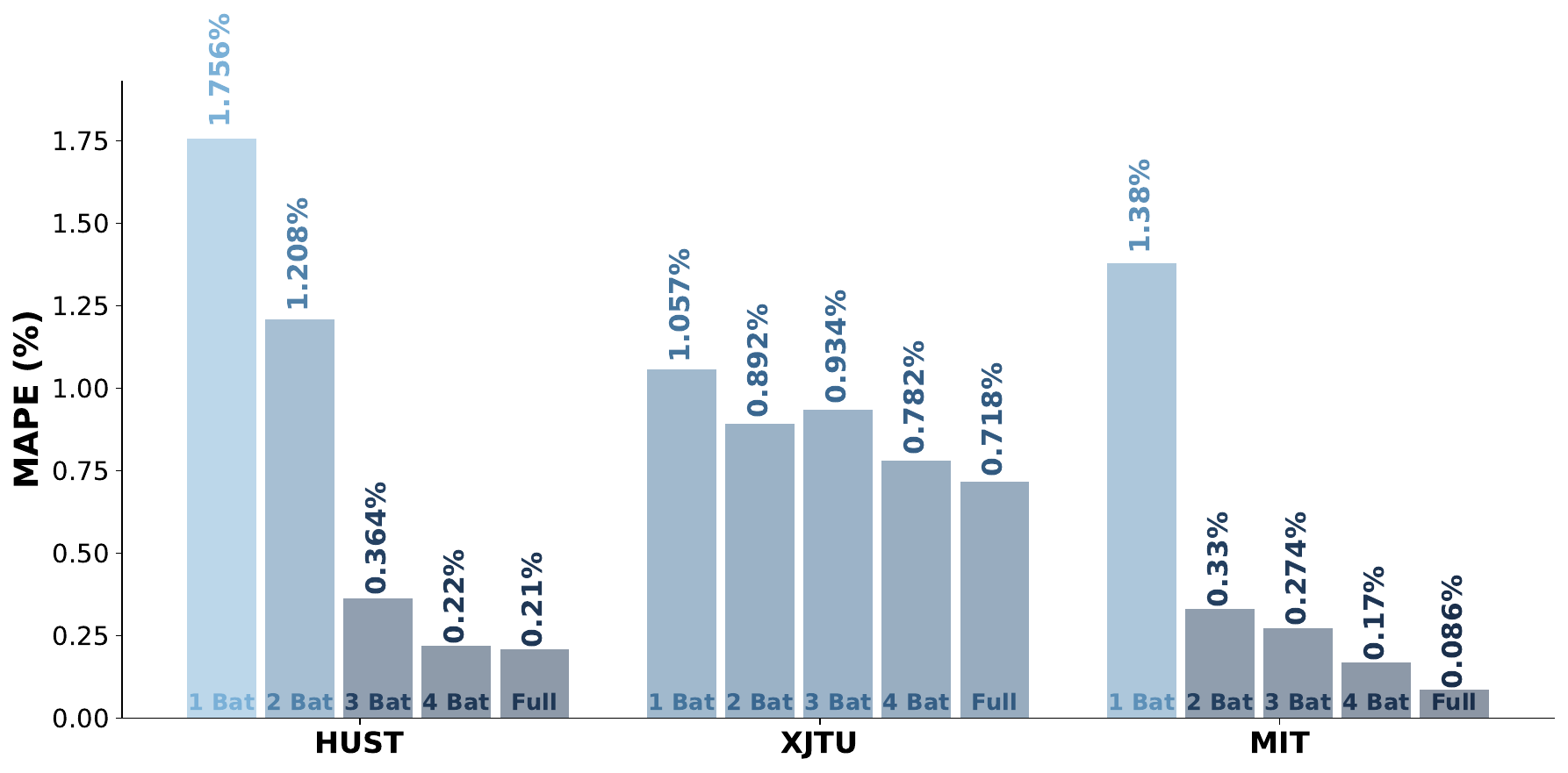}
    \caption{
        \textbf{Comparison of MAPE values under different training samples. }
        Models are trained with 1-4 batteries respectively and tested on the test set. The ``Full'' represents model performance when trained on the entire available dataset.
    }
    \label{fig:small_sample}
\end{figure}

Data-driven SOH models often require many batteries for stable training \cite{kraljevski2023machinelearningsmalldata}. We therefore train TC-SOH with only 1--4 batteries. Across XJTU, HUST, and MIT (Fig.~\ref{fig:small_sample}), four batteries already approximate full-data performance, indicating that temporal-contrastive representations reduce data dependence.

\vspace{-2mm}
\subsubsection{Cross-Domain Adaptation Ladder}

We evaluate cross-domain adaptability with a compact five-rung ladder: \textit{source-only}, \textit{FT-1bat}, \textit{FT-2bat}, \textit{Mix-1bat}, and \textit{Mix-2bat}. The TC module $g(\cdot)$ is frozen during transfer, while the encoder $f(\cdot)$ is adapted with one or two target batteries or jointly trained with source--target mixtures.

\begin{figure}[t]
    \vspace{-2mm}
    \centering
    \includegraphics[width=\linewidth, trim=0 5 0 5, clip]{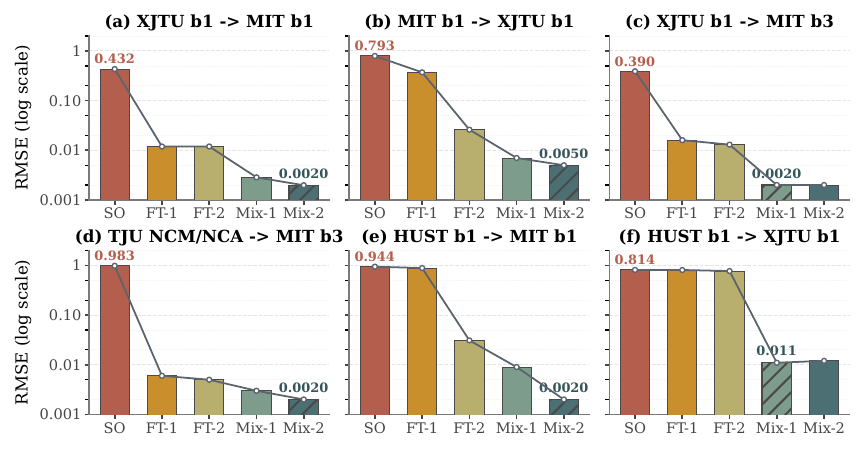}
    \vspace{-3mm}
    \caption{\textbf{Adaptation ladder.} RMSE (log scale) across strategies. Source-only fails catastrophically; a single target battery recovers performance; mixture training achieves the lowest errors.}
    \label{fig:adaptation_ladder}
\end{figure}

Fig.~\ref{fig:adaptation_ladder} shows that source-only prediction fails under large domain shifts (RMSE $\geq$ 0.39), while one or two target batteries reduce errors by one to two orders of magnitude. Mixture training is most stable, reducing all six transfer pairs below 0.012 RMSE, which suggests rapid adaptation when the encoder learns source--target structure jointly.

\subsection{Representation Analysis}

We organize the representation analysis as a progressive diagnostic chain. Visualization first checks whether embeddings form smooth degradation trajectories; sensitivity then asks whether they remain responsive in healthy and early-decay regimes where SOH changes are weak. Redundancy and probing analyses test whether this gain merely repackages hand-crafted descriptors or also contains residual SOH-relevant information. Finally, future-SOH probing and temporal shuffling evaluate whether the history-conditioned context $c_t=g(z_{\leq t})$ predicts later SOH through ordered cycle history rather than current-cycle readout alone. All reported results derive from the XJTU dataset (subset 1) under fixed charge-discharge protocols.

\begin{figure}[htbp]
    \vspace{-4mm}
    \centering

    \begin{minipage}[b]{0.40\linewidth}
        \centering
        \includegraphics[width=\linewidth, trim=10 10 10 10, clip]{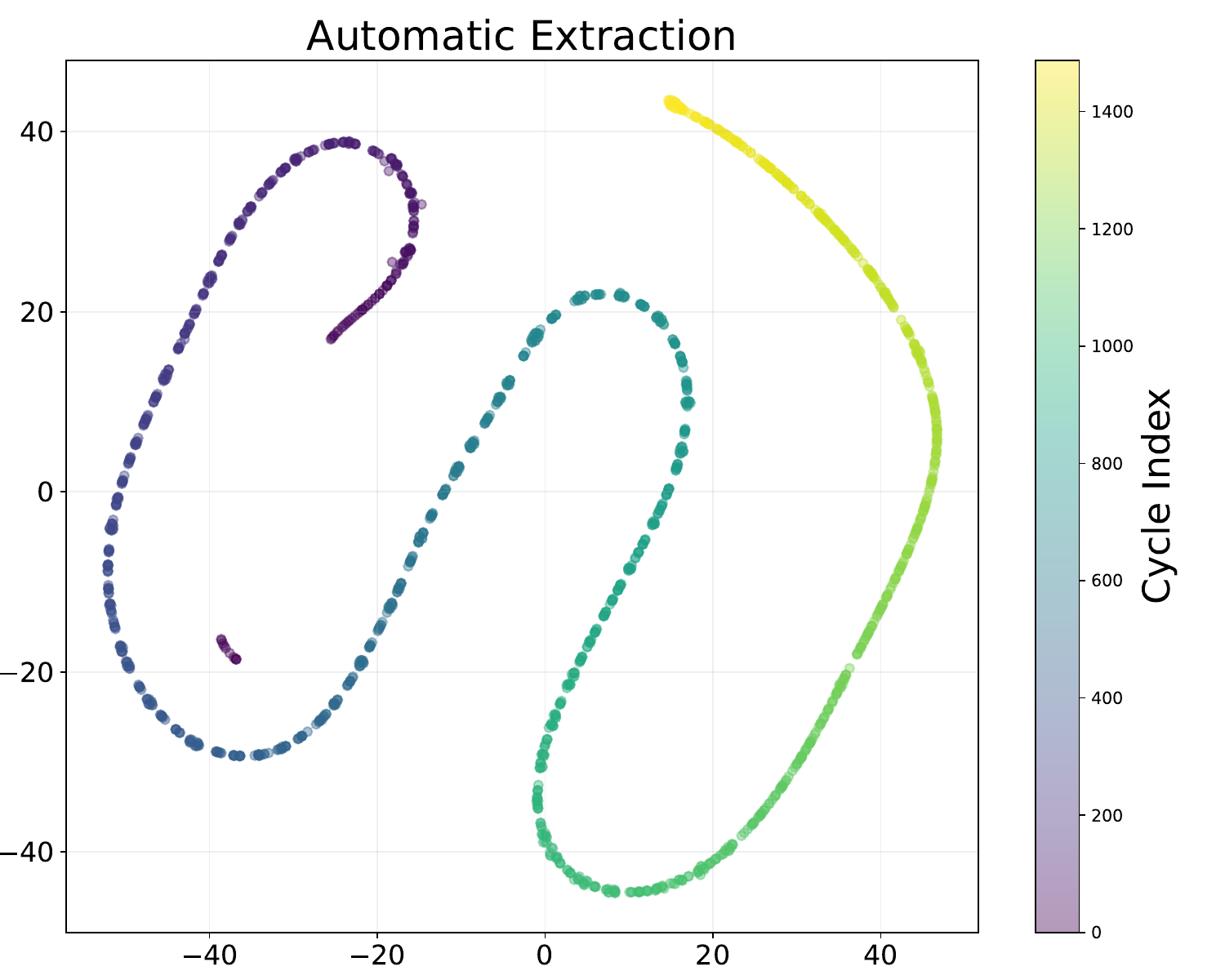}
        \vspace{-3mm} 
        \centerline{\footnotesize (a) TC-SOH}
    \end{minipage}
    \hfill 
    %
    \begin{minipage}[b]{0.40\linewidth}
        \centering
        \includegraphics[width=\linewidth, trim=10 10 10 10, clip]{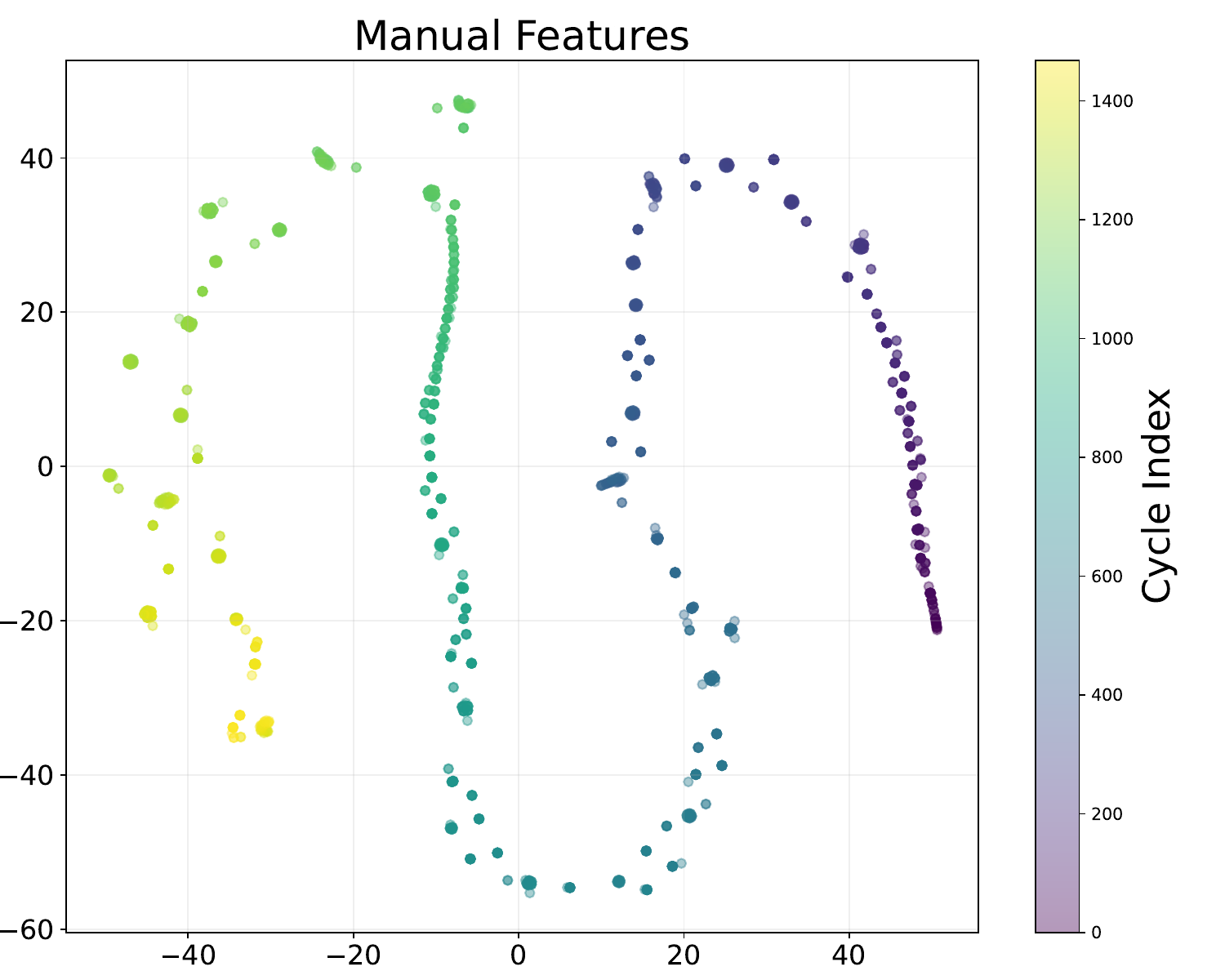}
        \vspace{-3mm}
        \centerline{\footnotesize (b) Manual Features}
    \end{minipage}
    \caption{
    \textbf{t-SNE Visualization on HUST.} Comparison between (a) automatically learned representations and (b) manual features.}
    \label{fig:t-sne_compare}

    \vspace{-5mm}
\end{figure}

\subsubsection{Feature visualization}
We use t-SNE\cite{JMLR:v9:vandermaaten08a} to project $(cycles,256)$ embeddings. Fig.~\ref{fig:t-sne_compare} shows smoother degradation trajectories for TC-SOH than for manual features, indicating stronger temporal continuity in the learned representation. This is the entry point of the diagnostic chain: adjacent cycles should move coherently as the cell ages rather than producing unstable pointwise embeddings; the next check asks whether this continuity also reflects weak early degradation.

\subsubsection{Sensitivity Analysis}
As an early-stage sensitivity check, we divide cycles into healthy (SOH $>0.95$), early decay ($0.90\leq$ SOH $\leq0.95$), middle decay ($0.80\leq$ SOH $<0.90$), and late/failure (SOH $<0.80$) regimes. In the nearly flat healthy regime, the same MLP predictor drops from $7.87\times10^{-5}$ MSE with manual features to $1.06\times10^{-5}$ with learned features; the CNN readout further reaches $3.50\times10^{-6}$. A similar advantage in early decay suggests that TC-SOH captures weak degradation cues before severe SOH drops dominate the signal. This narrows the interpretation from merely smooth trajectories to measurable early-stage degradation sensitivity.


\vspace{-2mm}
\subsubsection{Redundancy Analysis}
A simple explanation for the sensitivity result is that the encoder merely reproduces selected manual descriptors in smoother form. We therefore fuse manual descriptors with latent features using concatenation, Gated Fusion, SE-Block~\cite{hu2018squeeze}, and DepCA~\cite{kim2024cafo}. The deep-only stream remains strongest, while all fusion variants and the physics-only stream are weaker. Thus, the selected descriptors are largely redundant with TC-SOH representations: useful physically, but not complementary to the learned stream. This rules out feature fusion as the source of the main gain and motivates a more direct feature-space probe.

\begin{table}[t]
    \caption{Ablation study results of different fusion strategies.}
    \label{tab:ablation_study}
    \centering
    \renewcommand{\arraystretch}{1.05}
    \setlength{\tabcolsep}{3.5pt}
    \scriptsize
    \resizebox{\linewidth}{!}{
    \begin{tabular}{lcccc@{\hspace{1.5em}}lcccc}
        \toprule
        \multirow{2}{*}{Method} & \multicolumn{2}{c}{w/o Ortho} & \multicolumn{2}{c}{w/ Ortho} & \multirow{2}{*}{Method} & \multicolumn{2}{c}{w/o Ortho} & \multicolumn{2}{c}{w/ Ortho} \\
        \cmidrule(lr){2-3} \cmidrule(lr){4-5} \cmidrule(lr){7-8} \cmidrule(lr){9-10}
         & MAPE & RMSE & MAPE & RMSE & & MAPE & RMSE & MAPE & RMSE \\
        \midrule
        Concat   & 0.514 & 0.0058 & 0.505 & 0.0057 & DepCA       & 0.657 & 0.0071 & 0.589 & 0.0062 \\
        Gated    & 0.348 & 0.0056 & 0.359 & 0.0055 & Physic Only & 0.609 & 0.0097 & --    & --     \\
        SE-Block & 0.353 & 0.0070 & 0.423 & 0.0083 & \textbf{Deep Only} & 0.289 & 0.0033 & \textbf{0.285} & \textbf{0.0032} \\
        \bottomrule
    \end{tabular}}
\end{table}

\subsubsection{Probing Analysis}
Redundancy analysis shows that manual descriptors do not improve the predictor, but not how the two feature spaces overlap. We therefore adapt diagnostic probing from NLP~\cite{conneau2018cram}. Frozen deep features reconstruct capacity and energy with $R^2>0.99$ and voltage variance with $R^2=0.88$, showing overlap with expert descriptors. Conversely, manual features recover the dominant latent component but leave residual variation in the secondary PLS component. Thus, the latent space is physically interpretable, yet not exhausted by selected hand-crafted indicators. The remaining question is temporal: whether TC uses the ordered history of such embeddings to support later SOH prediction.

\begin{table}[t]
    \vspace{-2mm}
    \caption{Probing analysis: forward/reverse probing and SOH estimation with different feature sets.}
    \label{tab:feature_analysis_combined}
    \centering
    \setlength{\tabcolsep}{3.2pt}
    \renewcommand{\arraystretch}{1.05}
    \scriptsize
    \resizebox{0.94\linewidth}{!}{
    \begin{tabular}{lc@{\hspace{1.2em}}lc@{\hspace{1.2em}}lccc}
        \toprule
        \multicolumn{4}{c}{\textbf{Probing Task ($R^2$)}} & \multicolumn{4}{c}{\textbf{SOH Estimation}} \\
        \cmidrule(lr){1-4} \cmidrule(lr){5-8}
        \multicolumn{2}{l}{\textit{Deep $\to$ Physics}} & \multicolumn{2}{l}{\textit{Physics $\to$ Deep}} & Feature Set & Dim. & $R^2$ & RMSE \\
        \midrule
        Mean Volt.  & 0.999 & PLS Dim 1 & 0.974 & Raw Physics  & 5   & 0.985 & 0.0057 \\
        Energy      & 0.999 & PLS Dim 2 & \textbf{0.863} & \textbf{Full Deep} & 128 & \textbf{0.999} & \textbf{0.0016} \\
        CC Time     & 0.999 & --        & --    & PCA Dim 1    & 1   & 0.822 & 0.0199 \\
        Capacity    & 0.997 & --        & --    & PCA Dim 1-10 & 10  & 0.999 & 0.0018 \\
        Volt. Var.  & 0.875 & --        & --    & PLS Dim 1    & 1   & 0.986 & 0.0055 \\
        \bottomrule
    \end{tabular}}
    \vspace{-5mm}
\end{table}


\subsubsection{Future-SOH Probing and Temporal Shuffling}
The feature-level analyses leave one final question: whether the TC module organizes cycle embeddings into a state that predicts beyond current-SOH fitting. Since $c_t$ already aggregates hidden representations from cycles $1$ to $t$, we freeze the trained encoder--TC stack and train small ridge probes $\hat{y}_{t+k}=q(r_t)$ for $k\in\{1,5,10,20\}$, comparing manual features, $z_t=f(x_t)$, ablated contexts, and $c_t=g(z_{\leq t})$. This is stricter than current-SOH probing: because SOH is temporally smooth, a representation can fit the present cycle while failing to encode the degradation state needed several cycles ahead.

Temporal shuffling/reversal before $g(\cdot)$ provides the order test: if unordered feature statistics were sufficient, destroying cycle order would preserve most of the signal; degradation under shuffled or reversed prefixes indicates that TC uses the direction and accumulation of history.

\begin{table}[h!]
    \vspace{-1mm}
    \caption{Future-SOH probing on held-out cycles. Values are RMSE for probes trained on frozen representations.}
    \label{tab:future_soh_probing}
    \centering
    \setlength{\tabcolsep}{4.2pt}
    \renewcommand{\arraystretch}{1.05}
    \scriptsize
    \begin{tabular}{lcccc}
        \toprule
        Representation & SOH$_{t+1}$ & SOH$_{t+5}$ & SOH$_{t+10}$ & SOH$_{t+20}$ \\
        \midrule
        Manual features & 0.00747 & 0.00874 & 0.01029 & 0.01554 \\
        Single-cycle $z_t$ & 0.00506 & 0.00502 & 0.00556 & 0.00765 \\
        w/o TC context & 0.01370 & 0.00542 & 0.00619 & 0.00852 \\
        Shuffled context & 0.05078 & 0.05090 & 0.05063 & 0.05009 \\
        \textbf{TC context $c_t$} & \textbf{0.00439} & \textbf{0.00430} & \textbf{0.00467} & \textbf{0.00580} \\
        \bottomrule
    \end{tabular}
    \vspace{-3mm}
\end{table}

Table~\ref{tab:future_soh_probing} completes the diagnostic chain: TC context is the most predictive representation for future SOH, paired bootstrap tests confirm positive gains over $z_t$, and order-destroying controls weaken the probe. The gain therefore cannot be explained by smooth geometry, early sensitivity, or manual-feature recovery alone; it depends on the ordered temporal context learned by TC. In practical terms, the TC module contributes by compressing the prefix $z_{\leq t}$ into a degradation state that is both more informative than a single-cycle embedding and more useful than unordered temporal statistics. This is why the final claim is not simply that learned features are accurate, but that ordered temporal context provides additional information for subsequent SOH estimation.

\section{Conclusion} Manual feature pipelines for SOH estimation can become brittle under varying operating conditions\cite{10201510}. TC-SOH addresses this challenge with temporal-contrastive representation learning that extracts degradation-relevant features directly from charging data. Across 386 batteries from four public datasets, TC-SOH achieves competitive and often improved SOH estimation performance over the considered baselines, while also showing favorable robustness in transfer and low-data settings. Under a battery-level multi-seed evaluation protocol designed to reduce leakage risk\cite{GESLIN20231956}, these results support TC-SOH as a practical end-to-end alternative to hand-crafted feature design. Representation diagnostics further indicate that the learned features recover substantial information from selected physical descriptors while retaining additional SOH-relevant predictive variation, and that ordered temporal context improves subsequent-SOH prediction. Future work should connect these diagnostic findings with controlled electrochemical stress tests and broader baseline families.


\bibliographystyle{splncs04}
\bibliography{soh}

\end{document}